# Compressed Constraints in Probabilistic Logic and Their Revision

**Paul Snow**
Department of Computer Science
Plymouth State College
P.O. Box 6134
Concord, NH 03303 USA
paulsnow@oz.plymouth.edu

## Abstract

In probabilistic logic entailments, even moderate size problems can yield linear constraint systems with so many variables that exact methods are impractical. This difficulty can be remedied in many cases of interest by introducing a three-valued logic (true, false, and "don't care"). The three-valued approach allows the construction of "compressed" constraint systems which have the same solution sets as their two-valued counterparts, but which may involve dramatically fewer variables. Techniques to calculate point estimates for the posterior probabilities of entailed sentences are discussed.

## 1. PROLIFERATION OF WORLDS

An entailment problem in Nilsson's (1986) probabilistic logic derives an estimate for the prior probability of one sentence (hereafter, the "target") from the priors for a set of other ("source") sentences. The prior beliefs about the source sentences establish constraints of the form

$$P = VW$$
$$\Sigma \, w_i = 1 \qquad \text{sum over all "worlds"}$$
$$w_i \geq 0 \qquad \text{for all "worlds"}$$

Here, $P$ is the column vector of the sentences' priors. $V$ is a matrix derived from an inventory of all consistent patterns of truth assignments (1 = true, 0 = false) for the source and target sentences. For instance, for source sentences $Q$ and $Q => R$, and target sentence $R$, the consistent patterns (or "possible worlds") are the columns of the matrix:

| Q | 1 | 1 | 0 | 0 |
|------|---|---|---|---|
| Q=>R | 1 | 0 | 1 | 1 |
| R | 1 | 0 | 1 | 0 |

The matrix $V$ is the first two rows of the matrix above. The components $w_j$ of the vector $W$ are the unknown prior probabilities of the possible worlds, which are constrained by beliefs about the source sentences' priors.

The linear constraint system says that the prior of each sentence is the sum of the priors for the possible worlds in which the sentence is true. The system thus implies a constraint on the target sentence's prior. In the example, the prior for target sentence $R$ is $w_1 + w_3$. The linear system, by constraining $W$, also constrains this sum. The estimate for the target may either be a probability interval (computed using two linear programs), or else a point probability (perhaps maximizing entropy over the $w_j$'s).

A practical difficulty with this scheme is the large number of possible worlds that can arise with even a modest number of source sentences: ten sentences can yield a thousand worlds. Exponential complexity is inherent in the approach (Halpern, 1989).

One response to this difficulty is to introduce approximations, as Nilsson himself did. Kane (1989) suggested assessing conditional probabilities (rather than simply zero or one) in the entailment expression for the target sentence. Kane's method doesn't address proliferation of worlds on account of the source sentences, and requires additional assessments beyond priors for the source sentences.

The approach taken in this paper represents prior constraints without approximations or assessments beyond the source priors. Instead, a method is discussed that often derives smaller linear systems to convey constraints on the target sentence probability equivalent to those in Nilsson's original proposal. The effect should be that many moderate-sized entailment problems become practical for solution by exact means.

A method for revision which uses these compressed systems is also discussed. Because of the complicated interaction of conditionals and priors, simplifying assumptions are used.

## 2. OVERVIEW AND EXAMPLE

The method presented here derives linear constraint systems based on a three-valued logic common in engineering work (true, false, and "don't care," hereafter "d.c."). The opportunity to use the d.c. value arises whenever two worlds' truth assignment vectors differ in only one component, one world having true and the other false. In the example of the last section, worlds 1 and 3 could form one world with assignments [ d.c., 1, 1 ]. In effect, [ d.c., 1, 1 ] is a "shorthand" for the assertion that both [ 0, 1, 1 ] and [ 1, 1, 1 ] are possible worlds.

For example, consider a modus ponens with a conjunctive antecedent used by Kane (1989). The source sentences are independent A1, A2, A3, and the implication sentence A1 & A2 & A3 => B. The target is B. In this problem, there are 16 possible worlds:

|   |   |
|---|---|
| A1 | 1 1 0 0 0 0 0 0 0 0 1 1 1 1 1 1 |
| A2 | 1 1 0 0 0 0 1 1 1 1 0 0 0 0 1 1 |
| A3 | 1 1 0 0 1 1 0 0 1 1 0 0 1 1 0 0 |
| A1 & A2 & A3 => B | 1 0 1 1 1 1 1 1 1 1 1 1 1 1 1 1 |
| B | 1 0 1 0 1 0 1 0 1 0 1 0 1 0 1 0 |

With three-valued logic, these sixteen worlds can be compressed to five. The enumeration of these worlds follows readily from our understanding of the "conjunctive antecedent implies consequent" inference schema. An elementary understanding of what assignments are possible in that schema allows the specification of the possible worlds with $n$ conjuncts as follows:

1. one world where all sentences are true;
2. one world where all antecedents are true, and the implication and consequent are false;
3. for each i from 1 through n, a world where:
   conjunct #i is false
   conjuncts numbered < i are true
   conjuncts numbered > i are d.c.
   the implication is true
   the consequent is d.c.

Figure 1 shows a semantic tree for three antecedents constructed according to this plan. Figure 2 expresses the same information in a "matrix" format. The resulting five worlds for the example problem can inform a system of ordinary linear constraints. Where the original system contained equation constraints, the new system contains inequalities. Assuming that point priors are available for the $Pr(A_i)$'s, the specific system to bound $Pr(B)$ is (in

addition to the usual non-negativity and total probability constraints):

$$
\begin{aligned}
Pr(A1) &= [\, 1, 1, 1, 1, 0\,] \cdot W' \\
Pr(A2) &\geq [\, 1, 1, 1, 0, \mathbf{0}\,] \cdot W' \\
Pr(A2) &\leq [\, 1, 1, 1, 0, \mathbf{1}\,] \cdot W' \\
Pr(A3) &\geq [\, 1, 1, 0, \mathbf{0}, \mathbf{0}\,] \cdot W' \\
Pr(A3) &\leq [\, 1, 1, 0, \mathbf{1}, \mathbf{1}\,] \cdot W' \\
Pr(=>) &= [\, 1, 0, 1, 1, 1\,] \cdot W' \\
Pr(B) &\leq [\, 1, 0, \mathbf{1}, \mathbf{1}, \mathbf{1}\,] \cdot W' \\
Pr(B) &\geq [\, 1, 0, \mathbf{0}, \mathbf{0}, \mathbf{0}\,] \cdot W'
\end{aligned}
$$

(Values arising from d.c. assignments are in bold face; the operator "•" indicates the scalar product; W' is the transpose of W.) Discussion of the derivation of linear constraints from the semantic tree appears in section 4 below.

The possible solutions for $Pr(B)$ in the above system are identical to those that would be found by the corresponding two-valued system. Nevertheless, the "size" of the system (total number of vector or matrix components, a rough but fair estimator of the difficulty of the linear programs) is half that of the original.

## 3. COMPRESSION USING KNOWLEDGE AND USING SEARCH

Three-valued constraint systems are simplest to construct when they reflect a well-understood inference schema. The "conjunctive antecedent implies consequent" schema is, of course, at the heart of production rule-based systems. From this, the required constraint system can easily be constructed.

Based on the specification described in the last section, we see that $n$ antecedents yield only $n+2$ possible worlds. This compares with $2^{n+1}$ worlds in the two-valued system! (The total size of the linear system will be $O(n^2)$.)

In the case of arbitrary source and target sentences, tree construction will not be guided by an understanding of the possible truth assignments. Standard algorithms, like the Quine-McCluskey procedure, proceed by constructing the ordinary, two-value, semantic tree, and then search that tree for opportunities to combine worlds.

Note that compression may turn out to be impossible. Consider $m$ independent sentences $S_1 \ldots S_m$ as the source and the "parity function" ($S_1$ xor $S_2$ xor ... xor $S_m$) as the target. There are $2^m$ possible assignments, and each assignment has a Hamming distance of at least 2 from any other assignment. That is, no two assignments differ in only one component, and so no compression to use d.c. arises. In other cases, even if some compression were possible, the amount could be disappointing.

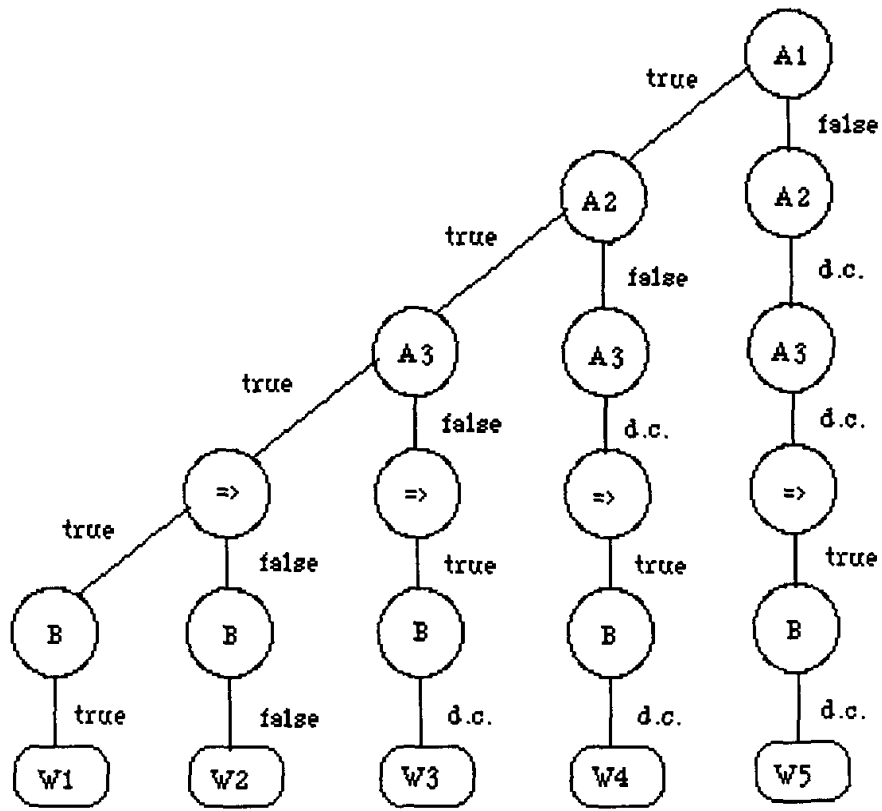

Figure 1. Compressed Semantic Tree for Kane (1989) Example.

|  | $W_1$ | $W_2$ | $W_3$ | $W_4$ | $W_5$ |
|---|---|---|---|---|---|
| $A_1$ | 1 | 1 | 1 | 1 | 0 |
| $A_2$ | 1 | 1 | 1 | 0 | d.c. |
| $A_3$ | 1 | 1 | 0 | d.c. | d.c. |
| $A_1A_2A_3 => B$ | 1 | 0 | 1 | 1 | 1 |
| B | 1 | 0 | d.c. | d.c. | d.c. |

Figure 2. Same Information as Fig. 1, in "Matrix" Format.

The effort involved can be considerable. Just constructing the two-value semantic tree is worst-case exponential, to say nothing of the search. Still, if extensive compression is achieved, the overall effort of the entailment problem may be reduced. That's because the linear programming steps are computationally intensive compared to tree construction, so the potential pay-off for finding a reasonable size constraint system can be handsome.

## 4. EXPRESSING THE CONSTRAINTS

Once the possible worlds are found, we can derive linear constraints. Returning to the Kane example, sentences like $A_1$ which have no d.c. values are handled just as they would be in Nilsson's proposal. For a sentence with a d.c. value, $A_2$, the semantic tree tells us that the possible assignments are [ 1, 1, 1, 0, d.c. ]. This can be read to mean that the prior probabilities $w_i$ of the worlds must be such that there is a number $p$ in the closed unit interval where

$$Pr(A_2) = w_1 + w_2 + w_3 + pw_5$$

This reading follows immediately from the way the fifth world was constructed. Of the total weight that was assigned to the original worlds that became world 5, some portion of that weight contributes to the sum that is $A_2$'s prior. For any $p$ and $W$ which satisfy the equation constraint just given, there is some apportionment of weight to the original worlds that achieves $p$ and which satisfies the two-valued system. Conversely, any solution of the two-valued system has a $p$ that satisfies the above equation.

The equation, in turn, is equivalent to two simultaneous inequality constraints

$$Pr(A_2) \leq w_1 + w_2 + w_3 + w_5$$
$$Pr(A_2) \geq w_1 + w_2 + w_3$$

A solution of the equation constraint for any admissible $p$ solves both inequalities, and for any solution of the inequalities, there is an admissible $p$ that satisfies the equation constraint.

The assignment for sentence $A_3$, [ 1, 1, 0, d.c., d.c. ], asserts the existence of two numbers (not necessarily

distinct) $q$ and $r$ in the closed unit interval, such that

$$Pr(A_3) = w_1 + w_2 + qw_4 + rw_5$$

Even though $r$ above and the earlier $p$ pertain to the same compressed world, their values are independent, a fact which follows from the way that the fifth world was constructed. Thus, the two constraints do not interfere with each other when they obtain simultaneously. The last equation can be translated into the inequalities

$$Pr(A_3) \leq w_1 + w_2 + w_4 + w_5$$
$$Pr(A_3) \geq w_1 + w_2$$

These simultaneous inequalities are equivalent to their equation constraint. Further, these inequalities can hold simultaneously with the inequalities derived earlier for the other sentence. (The full translation of this problem into simultaneous linear constraints was given near the end of section 2 above.)

To summarize, each equation constraint involving d.c. assignments can be expanded into two weak inequalities of opposite sense. The upper bounding sum includes the probabilities for the merged worlds; the lower bounding sum omits them. (For source sentences, the quantity being bound is the given prior for the sentence; for the target sentence, the bounds are on the unknown prior being sought).

Prior beliefs can be weak inequalities (for instance, to express bounds on the prior probability for a sentence). A weak inequality expression of belief in the original system yields one weak inequality of the same sense in the new system. If the inequality bounds a prior from above, then any merged worlds contribute to the sum; if from below, they do not.

## 5. REVISION WITH CONDITIONALS

If evidence $E$ is observed that bears on some sentence $S$, then we would wish to revise our probability estimates to reflect our new beliefs. Nilsson assumed that the effect of $E$ on any world $j$ depended only on whether $S$ was true in $j$ or not. In particular:

$$Pr(E \mid j) = Pr(E \mid S) \quad \text{if S is true in } j \quad (1a)$$
$$= Pr(E \mid \neg S) \quad \text{if S is false in } j \quad (1b)$$

(Nilsson made the assumption in a different, but equivalent, form: for sentences $S$, $T$, and evidence $E$ that bears on $S$, $Pr(T \mid S, E) = Pr(T \mid S)$ and $Pr(T \mid \neg S, E) = Pr(T \mid \neg S)$.) This assumption is frequently encountered in inference work. For a discussion of the motivation of the assumption in an A.I. context, see for example Pearl (1986). The assumption isn't strained, and is surely useful, but its main attraction is its simple relationship between the known conditionals given sentences and the unknown conditionals given worlds.

When $S$ is d.c. in $j$, we can extend this assumption by simple probabilistic identities to

$$Pr(E \mid j) = Pr(S \mid j) * Pr(E \mid S) + \\ Pr(\neg S \mid j) * Pr(E \mid \neg S) \quad (2)$$

By the way that the d.c. value was arrived at, knowledge that $j$ obtained would provide no constraint on the probability of $S$ that depends only on $j$ and $S$. This contrasts with the simpler situation when $S$ is either definitely true or definitely false in each world.

The sort of revision discussed in Nilsson's original proposal was to find a single feasible $W$ vector and then apply Bayes' formula to that prior. This yields a single point posterior for the distribution over worlds, and hence the target posterior. Another useful posterior constraint system describes all posterior distributions consistent with the prior constraints and assumption (1). This kind of estimate can be obtained from the uncompressed system by the straightforward application of a procedure for revising a linear prior system by a point conditional (Snow, 1991). The compressed system, unfortunately, generally doesn't easily support this kind of estimation. The conditional that comports with assumption (1) depends on the prior to which the conditional would be applied.

Revision using a chosen prior solution is not impeded by the interaction of conditionals and priors. The steps are:

- Choose a representative prior which solves the compressed system.

- Use that prior and the tableau ( e.g. figure 2 ) to assess feasible $Pr(S \mid j)$ values to replace d.c.'s among the sentences about which evidence may be seen.

- When evidence is observed, use equation (2) and the $Pr(S \mid j)$ values to compute a consistent conditional distribution over worlds.

- Apply Bayes rule using the chosen prior and the consistent conditional to compute the posterior estimate.

## 6. AN EXAMPLE OF REVISION

A complete inference problem would include estimates for the prior probability of each of the source sentences. Typically, the resulting prior constraint system will have many solutions for $W$, the possible prior distribution over the (compressed) worlds. The method described here doesn't depend on how the analyst selects which solution will serve as the representative prior over the worlds. Possibilities include the maximum entropy solution, or a solution which yields a prior for the target in the middle of its interval of possible values.

Suppose the estimated priors for the source sentences are:

$$Pr(A_1) = 0.8$$
$$Pr(A_2) = 0.7$$
$$Pr(A_3) = 0.6$$
$$Pr(=>) = 0.8$$

and the analyst chooses from among the solutions for the distributions over the worlds the representative solution

$$(0.2, 0.2, 0.2, 0.2, 0.2)$$

This choice is consistent with any estimate for the prior of $B$ in the range [ 0.2, 0.8 ]. The next step is replace the d.c. markers in figure 2 with estimated values of $Pr(S \mid j)$. Expecting to view evidence bearing on the source sentences, the analyst replaces their d.c.'s with

$$Pr(A_2 \mid 5) = 0.5$$
$$Pr(A_3 \mid 4) = 0.5$$
$$Pr(A_3 \mid 5) = 0.5$$

These choices aren't unique. They are consistent in the sense that

$$Pr(A_2) = w_1 + w_2 + w_3 + Pr(A_2 \mid 5) w_5$$
$$\text{and} \quad 0.7 = 0.2 + 0.2 + 0.2 + 0.5 * 0.2$$

Now suppose evidence $E$ is observed that bears on sentence $A_3$, and suppose that

$$Pr(E \mid A_3) = 0.8 \quad Pr(E \mid \neg A_3) = 0.4$$

Applying (1) and (2) gives as world conditionals

$$\text{Pr}(E\,|\,1) = 0.8$$
$$\text{Pr}(E\,|\,2) = 0.8$$
$$\text{Pr}(E\,|\,3) = 0.4$$
$$\text{Pr}(E\,|\,4) = 0.6$$
$$\text{Pr}(E\,|\,5) = 0.6$$

When these conditionals are applied to the representative prior, the corresponding posterior works out to

( 0.25, 0.25, 0.12, 0.19, 0.19 )

and the posterior estimate for **B** can be anything in the interval [ 0.25, 0.75 ].

Note that if evidence were observed that bore directly on **B**, the analyst would also have to assess **Pr( B | j )** for each of the three d.c.'s pertaining to **B**. This, of course, is just what Kane suggested. In that case, there would be a specific point estimate for the entailed posterior of the target, rather than an interval.

## 7. REVISION USING POSTERIORS

In his original proposal, Nilsson considered revision using posterior probabilities for sentences. The methods developed here for conditional revision can be applied directly to the case where a posterior for sentence **S** given evidence **E** rather than a conditional is known.

Assuming that neither **Pr( S )**, **Pr( ¬S )**, nor **Pr( E )** is zero, then we have the probabilistic identities

$$\text{Pr}(E\,|\,S)\ /\text{Pr}(E) = \text{Pr}(S\,|\,E)\ /\text{Pr}(S)$$
$$\text{Pr}(E\,|\,\neg S)\,/\text{Pr}(E) = \text{Pr}(\neg S\,|\,E)\,/\text{Pr}(\neg S)$$

Since we are assumed to know the truth status of **S** in each world, we know (or can assess consistent values for) the priors and any needed **Pr( S | j )**'s. Since division of all conditionals by the same constant (i.e., **Pr( E )**) has no effect on the calculated posterior, we can use the posterior to prior ratios in place of the conditionals in Bayes' formula. If **S** is d.c. in world **j**, and assumption (1) holds, then the appropriate conditional ratio is the average of the posterior to prior ratios weighted by **Pr( S | j )** and its complement, **Pr( ¬S | j )**.

## 8. CONCLUSIONS

The introduction of a third logical value can greatly reduce the number of variables seen in linear constraint systems for entailment in probabilistic logic. The method is most useful when the inference problem involves some easy-to-analyze inference schema, thus avoiding search in the construction of a semantic tree. Modus ponens with a conjunctive antecedent was emphasized here because of its importance in rule-based inference. Similar points could have been made using other common schemata involving modus tollens, modus tollendo ponens and other Latin friends for illustration.

For arbitrary entailments, a resort to search may be needed. The method in that case is something of a gamble: search costs can be heavy, and there is no guarantee that a substantial savings will be realized. Nevertheless, the gamble may be attractive in moderate sized problems which tax the means of linear programming. If successful, the analyst will trade many simple steps to avoid an impractical linear program.

The extra effort of applying the new method is substantially confined to the creation of a suitable three-value semantic tree. The derivation of a linear constraint system from the tree is conceptually simple, and no worse than twice as hard as deriving a two-value system from a conventional semantic tree with the same number of worlds.

Constraints obtained from three-valued systems, like other linear constraint systems, can be revised in the face of evidence using Bayesian approaches. Calculation of a point posterior from a chosen feasible prior is straightforward. Compressed systems do not appear to allow the conceptually easy calculation of an exact posterior constraint system because of the interaction between conditionals and priors. Uncompressed systems do have this capability. This difference between the two kinds of prior constraints is of limited practical import, since the size of uncompressed systems effectively precludes the actual use of the exact posterior system except in small problems.